\documentclass[journal]{IEEEtran}
\usepackage[utf8]{inputenc} % allow utf-8 input
\usepackage[T1]{fontenc}    % use 8-bit T1 fonts
\usepackage{hyperref}       % hyperlinks
\usepackage{url}            % simple URL typesetting
\usepackage{booktabs}       % professional-quality tables
\usepackage{amsfonts}       % blackboard math symbols
\usepackage{nicefrac}       % compact symbols for 1/2, etc.
\usepackage{microtype}      % microtypography
\usepackage{graphicx}
\usepackage{amssymb,amsmath,amsthm,array,amsfonts}
\usepackage[dvipsnames]{xcolor}
\usepackage{booktabs}
\usepackage{underoverlap}
\usepackage{subfigure}
\usepackage{multirow} 
\usepackage[rightcaption]{sidecap}
\sidecaptionvpos{figure}{c}
\usepackage{balance}
\usepackage[ruled,vlined,linesnumbered]{algorithm2e}
\usepackage{pifont}% http://ctan.org/pkg/pifont
\newcommand{\cmark}{\ding{51}}%
\newcommand{\xmark}{\ding{55}}%
\renewcommand{\t}[1]{\tiny{#1}}
\PassOptionsToPackage{hyphens}{url}\usepackage{hyperref} % to break url

\title{Ensemble Conformalized Quantile Regression for Probabilistic Time Series Forecasting}

\author{Vilde Jensen,
        Filippo Maria Bianchi$^{*}$,
        Stian Normann Anfinsen %~\IEEEmembership{Member,~IEEE}% <-this % stops a space
\thanks{Correspondence: *filippo.m.bianchi@uit.no}% <-this % stops a space
\thanks{V.\ Jensen did this work at Department of Physics and Technology, UiT The Arctic University of Norway, and is now with Kongsberg Satellite Services.}% <-this % stops a space
\thanks{F.\ M.\ Bianchi is with the Department of Mathematics and Statistics, UiT The Arctic University of Norway and NORCE Norwegian Research Centre.}% <-this % stops a space
\thanks{S.\ N.\ Anfinsen is with NORCE Norwegian Research Centre and Department of Physics and Technology, UiT The Arctic University of Norway.}% <-this % stops a space
}

\begin{document}

\maketitle
\begin{abstract}
This paper presents a novel probabilistic forecasting method called \textit{ensemble conformalized quantile regression (EnCQR)}. 
EnCQR constructs distribution-free and approximately marginally valid prediction intervals (PIs), which are suitable for nonstationary and heteroscedastic time series data. 
EnCQR can be applied on top of a generic forecasting model, including deep learning architectures.
EnCQR exploits a bootstrap ensemble estimator, which enables the use of conformal predictors for time series by removing the requirement of data exchangeability. 
The ensemble learners are implemented as generic machine learning algorithms performing quantile regression, which allow the length of the PIs to adapt to local variability in the data. 
In the experiments, we predict time series characterized by a different amount of heteroscedasticity.
The results demonstrate that EnCQR outperforms models based only on quantile regression or conformal prediction, and it provides sharper, more informative, and valid PIs.        
\end{abstract}


%%%%%%%%%%%%%%%%%%%%%%%%%%%%%%%%%%%%
%%% INTRO
%%%%%%%%%%%%%%%%%%%%%%%%%%%%%%%%%%%%
\section{Introduction}

In real-world planning and decision-making processes, accurate time series forecasting is essential and it is often desirable to express the uncertainty in the predictions through a \textit{probabilistic} forecast~\cite{gneiting2014probabilistic}. 
The most common way of obtaining probabilistic forecasts is by generating prediction intervals (PIs), which provide an admissible range of values for future observations with a specified confidence level \cite{lawless2005frequentist}. 
PIs are termed \textit{valid} if the coverage probability of future observations matches the specified confidence level.
% Valid PIs are essential in high-risk situations. 
The PI width is governed by the confidence level, where more uncertain predictions produce wider intervals, but also by the performance of the underlying prediction algorithm. 
When PIs become overly wide, they are less informative and denote poor performance in the prediction model. Thus, probabilistic forecasts should yield PIs that are as narrow as possible, while ensuring the designed confidence level~\cite{gneiting2014probabilistic}.

In application domains such as energy analytics, time series often exhibit strong seasonal and heteroscedastic behavior, as the variance in the observations relate to the cyclic nature of the data~\cite{nowotarskiprob, bianchi2017recurrent}.  
This makes some time intervals harder to predict, e.g., the variation in energy consumption is typically higher and more volatile during the day than in the night. 
When the variability is lower, one can make more confident predictions with narrower PIs. 
However, methods that construct fixed-length PIs are not able to model such a variability in the uncertainty and assign the same uncertainty to all time steps, often resulting in overly conservative PIs~\cite{liu2020hybrid}. 
To obtain more informative PIs for heteroscedastic time series, the length of the PIs should adapt to the variability at each time step.
\vspace{0.5cm}

%Conformal prediction is an on-top probabilistic forecasting framework that constructs marginally valid PIs based on a similarity measurement termed \textit{conformity} \cite{shafer2008tutorialCP}. CP makes no hard distributional assumptions, besides observations being exchangeable. Despite this appeal, conformal predictors construct PIs with constant or weakly varying length \cite{romano2019conformalized}, and the exchangeability assumption, which states that the information provided by the observations is independent of the order in which the observations are collected, makes CP unsuitable for time series data. Another probabilistic forecasting method is \textit{quantile regression}, producing PIs adaptive to heteroscedastic data by estimating quantile functions directly from the data \cite{gasthaus2019probabilistic}. However, the actual coverage of the resulting PIs is not guaranteed to satisfy the designed coverage level for finite samples.

\textbf{Contributions.} We directly tackle the challenge of constructing adaptive and valid PIs for time series data by combining and leveraging the strengths of quantile regression (QR) and conformal prediction (CP). 

CP is a probabilistic forecasting technique that constructs valid PIs in finite samples without making any distributional assumptions besides observations being exchangeable, which makes it unsuitable for time series data.
To apply CP to time series data we rely on the \textit{leave-one-out} ensemble prediction theory~\cite{evgeniou2004leave}. 
In addition, by using ensemble learners that perform QR, we generate PIs that adapt to the local variability in the time series.
%Our work enables CP to be used for time series data by utilizing the \textit{leave-one-out} ensemble prediction theory introduced in the \textit{EnbPI} algorithm by \cite{xu2020conformal_dynamic}, and adapts to local variability by using ensemble learners performing quantile regression. 

The proposed method, called \textit{ensemble conformalized quantile regression} (EnCQR), is flexible, as it can be placed on top of any QR algorithm. 
In addition, EnCQR is distribution-free and it constructs approximately valid PIs independently of the accuracy of the underlying prediction method. 

{\color{blue}
\begin{table}[!ht]
\centering
\begin{tabular}{lcc}
                         & Adaptive PI & Valid PI \\ \hline
\multicolumn{1}{l|}{QR}  & \cmark      & \xmark   \\
\multicolumn{1}{l|}{CP}  & \xmark      & \cmark   \\
\multicolumn{1}{l|}{EnCQR} & \cmark      & \cmark  
\end{tabular}
\end{table}
}

We test the performance of EnCQR on five real-world datasets from application domains where it is common to encounter heteroscedastic time series. We show that EnCQR can produce valid and adaptive PIs for such data.
To demonstrate the versatility of EnCQR, we use it on top of three different regression algorithms: random forest regression and two neural networks for time series data. 
Results show that, compared to the state of the art, EnCQR generates narrower yet valid PIs. 

%%%%%%%%%%%%%%%%%%%%%%%%%%%%%%%%%%%%
%%% BACKGROUND
%%%%%%%%%%%%%%%%%%%%%%%%%%%%%%%%%%%%
\section{Background}

\subsection{Prediction Intervals (PIs)}
Let $X \in \mathcal{X}$ and $Y \in \mathcal{Y}$ be random variables representing the input observation and label, respectively.
We denote with $\pi(X,Y)$ the joint distribution of $X$ and $Y$, and with $\pi(Y|X)$ the conditional distribution of $Y$ given $X$.
A PI constructed using a collection of training samples $\{(x_i,y_i), i=1,2,\ldots,n\}$, where $(x_i,y_i)$ are realizations of $\pi(X,Y)$, is given by $\hat{C}_{\pi,n}(x) = [L(x),U(x)]$, where $L$ and $U$ are functions that map $\mathcal{X}$ into $\mathcal{Y}$. 
The width, or length, of a PI $U(X_{n+1}) - L(X_{n+1})$ is governed by the confidence level $\alpha$, i.e.\ the probability $(1-\alpha)$ of a new observation lying within the PI.
More uncertain predictions produce wider intervals.
The \textit{coverage} of a PI indicates the probability that the interval contains the actual value of the predicted variable.
A PI is called \textit{valid}, or calibrated, if the coverage probability for a new test point $(X_{n+1},Y_{n+1}) \sim \pi$ is guaranteed to be equal or greater than the designed confidence level.  
PI's coverage guarantees are discussed in the following. 

% ---------------------------------------
\subsection{Marginal and Conditional Coverage}
The PI's coverage guarantee can be defined on average over a set of test points (\textit{marginal} coverage guarantee) or pointwise for any fixed value $X_{n+1} = x$ (\textit{conditional} coverage guarantee)~\cite{barber2019marginal}. 
For a distribution-free marginal coverage guarantee, the probability that the PI covers the true test value $Y_{n+1}$ must be at least $1-\alpha$ on average over a random draw from any underlying distribution $\pi$:
\begin{equation}
    \mathbb{P}\left\{ Y_{n+1} \in \hat{C}_{\pi,n}(X_{n+1})\right\} \geq 1 - \alpha \,.
\end{equation}

Conditional coverage is a much stricter definition compared to marginal coverage and, hence, harder to ensure. 
A PI satisfies conditional coverage on the $1-\alpha$ level if
\begin{equation}
    \mathbb{P}\left\{ Y_{n+1} \in \hat{C}_{\pi,n}(X_{n+1})|X_{n+1} = x\right\} \geq 1 - \alpha\,
\end{equation}
meaning that for any point $x$, the probability that $\hat{C}_{\pi,n}$ covers $X_{n+1}=x$ must be at least $1-\alpha$. To demonstrate the difference between marginal and conditional coverage, \cite{barber2019marginal} presents the following example:
\begin{quote}
    \textit{Suppose that each data point i corresponds to a patient, with $X_i$ encoding relevant covariates (age, family history, current symptoms, etc.), while the response $Y_i$ measures a quantitative outcome (e.g., reduction in blood pressure after treatment with a drug). When a new patient arrives at the doctor’s office with covariate values $X_{n+1}$, the doctor would like to be able to predict their eventual outcome $Y_{n+1}$ with a range, making a statement along the lines of: “Based on your age, family history, and current symptoms, you can expect your blood pressure to go down by 10–15 mmHg”.}
\end{quote}
When setting $\alpha = 0.05$, the statement made by the doctor should hold with a probability of 95\%. 
For marginal coverage, the statement has a 95\% probability of being accurate on average for all possible patients. 
Since we consider the average, the statement might have a significantly lower, even 0\%, chance of being accurate for patients of a specific age group, but is compensated by a coverage probability that is higher than 95\% for the other age groups.
On the other hand, for conditional coverage the statement made by the doctor must hold with 95\% probability for every individual patient, regardless of age. Therefore, conditional coverage is more difficult to ensure.

Due to the stricter requirements, conditional coverage cannot be satisfied in distribution-free settings \cite{xu2020conformal_dynamic}. Consequently, most probabilistic forecasting methods focus on satisfying marginal coverage, or a compromise between marginal and conditional coverage. 
Please notice that methods ensuring marginal coverage can possibly, but not necessarily, obtain conditional coverage as well.

% ---------------------------------------
\subsection{Ensemble Learning}
Ensemble learning is a performance-enhancing technique for statistical and machine learning algorithms.
In ensemble learning, a prediction model is built by using a collection of simpler base models \cite{hastieelementsstatistical}, each one independently optimized to solve the same problem. 
In the context of machine learning, an ensemble model can be broadly defined as a system constructed with a set of individual models working in parallel and whose outputs are combined with a decision fusion strategy to produce a single answer for a given problem~\cite{huang2009ensemble}. 

By combining a group of weak learners into one strong/expert learner, ensemble learning produces significantly improved results compared to individual learners~\cite{kim2020jackknife}. 
The learners in the ensemble are often termed \textit{member} learners. They can be any machine learning algorithm, such as neural networks, support vector machines, or decision trees \cite{huang2009ensemble}. The ensemble of learners can be generated using three different approaches:
\begin{itemize}
    \item \textit{heterogeneous} ensembles: the member learners come from several different classification or regression algorithms;
    \item \textit{homogeneous} ensembles: member learners are generated using the same algorithm, but they are trained on different data;
    \item a combination of the two techniques.
\end{itemize}

The homogeneous ensemble method is formalized in Algorithm \ref{alg:ensemblelearning}. 
\IncMargin{1.5em}
\begin{algorithm}[!htpb]
  \DontPrintSemicolon
  \SetKwInOut{Input}{input}
  \SetKwInOut{Output}{output}

\Indm\Indmm
  \Input{Data $\{(x_i,y_i)\}_{i=1}^n$, base learning algorithm $A$, aggregation function $\phi$}
  \Output{Ensemble regression function $\hat{\mu}_\phi$}
\Indp\Indpp
  \BlankLine
  \For{$b = 1,\ldots, B$}{ 
    Sample an index set $S_b = (i_{b,1},\ldots, i_{b,m})$ from indices $(1,\ldots, n)$\ with or without replacement\;
    Compute $\hat{\mu}_b = A((X_{i_{b,1}},Y_{i_{b,1}}), \ldots, (X_{i_{b,m}},Y_{i_{b,m}})) $}
    Define $\hat{\mu}_\phi = \phi(\hat{\mu}_1,\ldots,\hat{\mu}_B)$
    \BlankLine
    \Indm\Indmm
\Indp\Indpp
\caption{Homogeneous Ensemble Learning}
\label{alg:ensemblelearning}
\end{algorithm}
\DecMargin{1.5em}
The ensemble method is implemented in two steps. 
First, a population of $B$ base learners is trained using different datasets, or different \textit{multisets} sampled from the available training data. 
Then, the predictions of the base learners are combined to form a single predictor by using an aggregation function $\phi$.
Popular aggregation functions include the mean, median, or trimmed mean. Using different aggregation functions have different benefits, e.g.\ the mean reduces the mean square error (MSE), the median reduces the sensitivity to outliers, and the trimmed mean gives a compromise of both \cite{xu2020conformal_dynamic}. 

The multisets can be obtained from the training data by using several different methods, e.g.\ \textit{bootstrapping} or \textit{subsampling} \cite{kim2020jackknife}. 
Bootstrapping creates multiple samples from the original data by randomly sampling \textit{with} replacement \cite{hastieelementsstatistical}, where the sampled multisets often have the same size as the original dataset. Contrarily to bootstrapping, subsampling creates multisets from the original data by extracting subsets \textit{without} replacement. 
One of the earliest and simplest ensemble method is bootstrap aggregating, or \textit{bagging} for short. 
Bagging uses bootstrapping together with the mean aggregation function to create ensemble models. 
% Bagging obtains diversity between member learners by training the individual learners on different samples of the same dataset. 
Subsample aggregating, or \textit{subagging}, is a variant of bagging, where subsampling replaces bootstrapping. 

Ensemble learners can produce very accurate results, since combining several models with relatively similar bias reduces the variance and improves the generalization capability \cite{zhang2012ensemblebook}. However, this is only the case if the member learners are sufficiently diverse and accurate. Diversity between member learners is essential for the ensemble performance, since little is gained by combining a vast ensemble of learners if they all produce the same result. 
%The core idea of ensemble learning is to combine a set of diverse models that together produce satisfactory results by compensating each other’s shortcomings. 
It is challenging to obtain diversity among the member learners since they are all optimized to solve the same task and, usually, they are trained on data derived from the same dataset, which makes the learners being highly correlated. 
There is a trade-off between the performance of the individual learners and the diversity among them. Combining very accurate but highly correlated learners often gives worse results than a combination of accurate and less accurate learners, since complementarity is more important than individual performance \cite{zhou2012ensemblefoundations}. 

% ---------------------------------------
\subsection{Conformal Prediction (CP)} 
CP is an on-top probabilistic forecasting framework that constructs marginally valid PIs based on a similarity measurement called \textit{conformity} \cite{shafer2008tutorialCP}. 
CP makes no hard distributional assumptions besides that observations must be exchangeable, i.e., the information provided by the observations is independent of the order in which the observations are presented. 
Despite this appealing feature, CP constructs PIs with constant or slightly varying length \cite{romano2019conformalized}.
In addition, the exchangeability assumption makes CP unsuitable for time series data.

CP was first introduced as a transductive inference method~\cite{gammerman1998}, where different data realizations are presented several times to the underlying learning algorithm, making it unsuitable for models that during training iterate through the data samples until convergence~\cite{kath2021conformal}. 
The \textit{inductive} conformal prediction, proposed by \cite{papadopoulos2002inductive}, avoids the shortcomings of the transductive CP method, but requires the training data to be split into two disjoint sets. 
From now on, we refer to the inductive method when mentioning CP. 

For regression problems, CP starts from a training set $\{(X_i,Y_i)\}_{i=1}^n$ with pairs of predictors $X_i\in\mathbb{R}^d$ and response variables $Y_i\in\mathbb{R}$, and splits it into two subsets: the proper training set, $\mathcal{I}_1$, and a calibration set, $\mathcal{I}_2$. A regression model is fitted using $\mathcal{I}_1$, while the conformity score -- a statistic of the prediction errors (residuals) obtained from $\mathcal{I}_2$ -- is used to quantify the uncertainty in future predictions.
Given a new observation $X_{n+1}=x$, we require that the conditional PI for the regressand $Y_{n+1}$ with miscoverage rate $\alpha$, denoted $\hat{C}_\alpha(x)$, must satisfy:
\begin{equation}
\mathbb{P}\left\{Y_{n+1}\in\hat{C}_\alpha(X_{n+1}=x)\right\}\geq 1-\alpha\,.
\end{equation}

CP provides this conditional PI as:
\begin{equation}
    \hat{C}_\alpha(x) = [\hat{\mu}(x) - Q_{1-\alpha}(\mathcal{R},\mathcal{I}_2),\hat{\mu}(x) + Q_{1-\alpha}(\mathcal{R},\mathcal{I}_2)]\,,
    \label{eq:cp_pi}
\end{equation}
where $\hat{\mu}(x)$ is the prediction made by the underlying regression model,
$\mathcal{R}$ is the set of residuals $R_i$ computed from the predictions of the samples $i\in\mathcal{I}_2$, and the conformity score $Q_{1-\alpha}(\mathcal{R},\mathcal{I}_2)$ is the $(1-\alpha)$-th quantile of $\mathcal{R}$~\cite{romano2019conformalized}. 
The residuals used to obtain the conformity score are often computed with the $L_1$ norm, but other distance measures can be used. From Eq.~\eqref{eq:cp_pi} it is clear that CP was designed with homoscedastic data in mind, since the PI is constructed as the conditional mean estimate of the response variable with a fixed-width band around it \cite{sesia2020comparison}. 

For an in-depth introduction to CP and its applications, we refer the interested reader to a recent tutorial~\cite{angelopoulos2021gentle}.

% ---------------------------------------
\subsection{Quantile Regression (QR)} 
QR aims at estimating a conditional quantile function (CQF) of $Y$ given $X$ at the specified $\alpha$. The CQF is defined as:
\begin{equation}
q_\alpha(x)=\inf\{y\in\mathbb{R}:F_Y(y|X=x)\geq\alpha\}\,,    
\end{equation}
where $F_Y(y)$ is the conditional distribution function of $Y$, whose probability density function can be estimated from empirical CQFs with miscoverage in the range $(0<\alpha<1)$~\cite{Koenkerquantile,taylor2000quantile}. 
PIs can be obtained directly from two empirical CQFs computed from the training set.
The confidence level $(1-\alpha)$ of the PI is the difference between such two quantile levels. 
The estimated conditional PI of QR thus becomes:
\begin{equation}
    \hat{C}_\alpha(x) = \big[ \hat{q}_{\alpha_{lo}}(x),\hat{q}_{\alpha_{hi}}(x) \big],
    \label{eq:conditionalpredinterval}
\end{equation}
where $\hat{q}_{\alpha_{lo}}(x)$ and $\hat{q}_{\alpha_{hi}}(x)$ are the empirical CQFs computed for $\alpha_{lo} = \alpha/2$ and $\alpha_{hi} = 1 - \alpha/2$. 

Unlike the PI in Eq.~\eqref{eq:cp_pi}, the width of the PI in Eq.~\eqref{eq:conditionalpredinterval} depends on each specific data point $x$ and can vary significantly from point to point. Therefore, QR yields intervals that adapt to heteroscedasticity in the data. 
However, when the ideal interval $C_\alpha(x)$ is replaced by the finite sample estimate $\hat{C}_\alpha(x)$ in Eq.~\eqref{eq:conditionalpredinterval}, the actual coverage of the PI is not guaranteed to match the designed confidence level $(1-\alpha)$~\cite{romano2019conformalized}.

The estimation of $\hat{q}_{\alpha_{lo}}(x)$ and $\hat{q}_{\alpha_{hi}}(x)$ can be cast as a optimization problem that minimizes the \textit{pinball loss}. The pinball loss of an observation pair $(x_i,y_i)$ is defined as: 
\begin{equation}
    L_{\alpha,i} = 
    \begin{cases}
    (1-\alpha)(\hat{q}_\alpha(x_i)-y_i), & \hat{q}_\alpha(x_i) \geq y_i\,; \\
    \alpha(y_i - \hat{q}_\alpha(x_i)), & \hat{q}_\alpha(x_i) < y_i\,,
    \end{cases}
    \label{eq:pinball}
\end{equation}
where $y_i$ denotes the $i$-th sample response and  $\hat{q}_\alpha(x_i)$ is the $\alpha$-th quantile estimated for the corresponding predictor~\cite{wang2019probabilisticlstmpinball}. The pinball loss measures how well the estimated quantile relates to the actual distribution of the data: the lower the pinball loss, the more accurate is the estimation.
The pinball loss can be used as the objective function to train a deep learning model~\cite{smyl2020hybrid, dudek2021hybrid, smyl2022esdrnn}.
We refer to a neural network (NN) that performs QR and is optimized with Eq.\ \eqref{eq:pinball} as a \textit{QRNN}.

% ---------------------------------------
\subsection{Conformalized quantile regression (CQR)} 
CQR is a probabilistic forecasting method that combines CP and QR to construct valid PIs for heteroscedastic data~\cite{romano2019conformalized, kivaranovic2020adaptive}. 
CQR inherits the advantages of both QR and CP: the properties of QR allow the method to adapt to the local variability in the data and the use of CP guarantees valid marginal coverage. 
Similarly to CP, CQR assumes the samples to be exchangeable and splits the training data into a proper training set and a calibration set. 
The resulting PIs are conformalized using the conformity scores
\begin{equation}
E_i = \text{max}\left\{\hat{q}_{\alpha_{lo}}(x_i)-y_i,y_i - \hat{q}_{\alpha_{hi}}(x_i)\right\},\;i\in\mathcal{I}_2
\label{eq:cqr_conf}
\end{equation}
which quantify the error made by the PI of the stand-alone QR algorithm, as specified by the bounds $\hat{q}_{\alpha_{lo}}(x_i)$ and $\hat{q}_{\alpha_{hi}}(x_i)$ in Eq.\ \eqref{eq:conditionalpredinterval}. 

The CQR PIs are calculated as follows:
\begin{equation}
    \begin{split}
    \hat{C}_\alpha(x) = \left[\right. &\hat{q}_{\alpha_{lo}}(x) - Q_{1-\alpha}(\mathcal{E},\mathcal{I}_2),\\
    &\hat{q}_{\alpha_{hi}}(x) + Q_{1-\alpha}(\mathcal{E},\mathcal{I}_2)\left.\right]\,,
    \end{split}
\end{equation}
where $\mathcal{E}=\{E_i\}_{i\in\mathcal{I}_2}$.  Note that the value of $Q_{1-\alpha}(\mathcal{E},\mathcal{I}_2)$, which is used to conformalize the PIs constructed by the QR algorithm, is fixed for all new data points $x$, similarly to $Q_{1-\alpha}(\mathcal{R},\mathcal{I}_2)$ in CP.

The use of CQR in combination with QRNNs can yield unnecessarily wide PIs \cite{romano2019conformalized}. 
This problem can be addressed by tuning the nominal quantile levels, $\alpha_{lo}$ and $\alpha_{hi}$, of the underlying QRNN as additional hyperparameters, which does not invalidate the coverage guarantee. 

% ---------------------------------------
\subsection{Ensemble Batch Prediction Intervals (EnbPI)} 
EnbPI is a method inspired by CP that builds distribution-free PIs for nonstationary time series~\cite{xu2020conformal_dynamic}. 
EnbPI assumes a time series data generating process on the form:
$$
Y_i = f(X_i) + \epsilon_i, \quad i = 1,2,3,\dots
$$
The error process $\{\epsilon_i\}_{i\geq1}$ is assumed to be stationary and strongly mixing, which replaces the exchangeability assumption required by CP. 
The probabilistic forecasts are constructed by aggregating point forecasts produced from leave-one-out predictions\footnote{When dealing with time series data, future data points are those left outside (i.e., leave-future-out).} constructed using homogeneous bootstrapped ensemble estimators.
The diversity between the ensemble learners is obtained by training them on different subsets of the original training dataset. 

The aggregated point predictions are used to build a PI with width equal to the $(1-\alpha)$-th empirical quantile of the latest $T$ observed residuals. 
Given the training data $\{(x_i,y_i)\}_{i=1}^T$, the PI at time $t$ is defined as:
\begin{equation}
\begin{split}
    \hat{C}_{\alpha}(x_t) = \left[\right.
    &\hat{f}_{-t}(x_t) - Q_{1-\alpha}(\epsilon_T),\\
    &\hat{f}_{-t}(x_t) + Q_{1-\alpha}(\epsilon_T)\left.\right]\,,
    \label{eq:enbpiPI}
\end{split}
\end{equation}
where $\epsilon_T=\{\hat{\epsilon}_i\}_{i=t-1}^{t-T}$. 
The residuals used to obtain the conformity score are computed as the absolute error between the training sample responses and the predictions of the leave-one-out estimators, denoted $\hat{f}_{-t}(x_t)$.

To handle nonstationary time series data, EnbPI exploits a sliding window of size $s$ to account for new data without refitting the underlying regression algorithm. 
In particular, the list containing the $T$ out-of-sample residuals is updated every $s$ predictions.
This allows the width of the subsequent PIs to vary and makes the calibration of the PI widths more dynamic and accurate. 
The ensemble learners are only trained once and are used to predict the center of the PIs for the future time steps. 
Hence, the ensemble learners are assumed to model $f$ sufficiently well. 
In practice, this assumption can fail when the window size $s$ is large and long-term predictions are made. 
Indeed, if the dynamics of nonstationary time series significantly change over time, the original model $f$ will eventually stop to describe well the underlying process. 
Valid coverage can still be obtained if a small window size is used, but the resulting intervals can become inflated if the out-of-sample absolute residuals are large.

%%%%%%%%%%%%%%%%%%%%%%%%%%%%%%%%%%%%
%%% METHOD
%%%%%%%%%%%%%%%%%%%%%%%%%%%%%%%%%%%%
\section{The EnCQR Algorithm}

The proposed \textit{ensemble conformalized quantile regression} (EnCQR) algorithm, summarized in Alg.~\ref{alg:enCQR}, combines an ensemble of QR learners with CP to construct PIs for time series. 
EnCQR consists of three main steps: 
\begin{enumerate}
    \item \textbf{Train the ensemble learners (lines 1-9)}. 
    The homogeneous learners are trained on independent subsets constructed in lines 1-4.
    Next, the ensemble learners are used to construct leave-one-out estimates for each observation $i$, by aggregating all the learners trained on subsets not including sample $i$ (line 7). Then, the conformity scores between the aggregated leave-one-out predictions and the training labels are computed (line 8). 
    
    \item \textbf{Sequentially construct PI for the observations in the test set (lines 10-15)}. The observations in the test set are predicted using the ensemble learners, which produce a set of $B$  quantile functions for both the upper and lower PI limit.
    The final PI limits are obtained by first aggregating the estimated quantile functions and then by conformalizing them using the $(1-\alpha)$-th quantile of the out-of-sample residuals calculated during training.
    
    \item \textbf{Update the residuals (lines 16-20)}. The out-of-sample residuals are updated after every $s$ new observations are predicted by replacing the oldest $s$ elements of the list, such that its length remains the same. 
\end{enumerate}

\IncMargin{1.2em}
\begin{algorithm}[!ht]

  \DontPrintSemicolon
  \SetKwInOut{Input}{input}
  \SetKwInOut{Output}{output}

\Indm\Indmm
  \Input{Training data $\{(x_i,y_i)\}_{i=1}^T$, quantile regression algorithm $A$, confidence level $\alpha \in (0,1)$, aggregation function $\phi$, no. of ensemble models $B$, window size $s$, and test data $\{(x_i,y_i)\}_{i=T+1}^{T+T'}$, where $y_i, i=T+1,\ldots,T+T'$ become available only after the batch of $s$ PIs is constructed.}
  \Output{Ensemble PIs $\{\hat{C}_{\alpha,T}(x_i)\}_{i=T+1}^{T+T'}$}
\Indp\Indpp
  \BlankLine
    Determine length of ensemble subset: $T_b \leq \nicefrac{T}{B}$\;  \For{$b = 1,\ldots, B$}{ 
    Sample index set $S_b = (i_{T_b \times b},\ldots, i_{T_b \times b + T_b})$ from indices $(1,\ldots, T)$\;
    Fit quantile ensemble estimator $[\hat{q}_{\alpha_{lo}}^{(b)},\hat{q}_{\alpha_{hi}}^{(b)}] = A( \{ (x_i,y_i) \}_{i \in S_b})$}
    Initialize $\mathcal{E}_{lo} = \{\}$ , $\mathcal{E}_{hi} = \{\}$\;
    
  \For{$i = 1,\ldots, T$}{ 
    $\left[\hat{q}_{\alpha_{lo}}(x_i), \hat{q}_{\alpha_{hi}}(x_i)\right] = [ \phi(\{\hat{q}_{\alpha_{lo}}^{(b)}(x_i)\}_{b \in \mathcal{B}_{-i}}), \phi(\{\hat{q}_{\alpha_{hi}}^{(b)}(x_i)\}_{b \in \mathcal{B}_{-i}})]$ with $\mathcal{B}_{-i} = \{S_b \;\text{s.t.}\; i \notin S_b \}$\;
    Compute $E_{lo_i}$ and $E_{hi_i}$ as in Eq. (\ref{eq:assym_nc}) \;
    $\mathcal{E}_{lo} = \mathcal{E}_{lo} \cup \{E_{lo_i}\}$,
    $\mathcal{E}_{hi} = \mathcal{E}_{hi} \cup \{E_{hi_i}\}$}
    
  \For{$i = T+1,\ldots, T + T'$}{ 
    Let $\hat{q}_{\alpha_{lo}}(x_i) = \phi(\{\hat{q}_{\alpha_{lo}}^{(b)}(x_i)\}_{b=1}^B)$\;
    Let $\hat{q}_{\alpha_{hi}}(x_i) = \phi(\{\hat{q}_{\alpha_{hi}}^{(b)}(x_i)\}_{b=1}^B)$\;
    Let $\omega_{lo_i} = (1-\alpha)$-th quantile of $\mathcal{E}_{lo}$\; 
    Let $\omega_{hi_i} = (1-\alpha)$-th quantile of $\mathcal{E}_{hi}$\; 
    Return $\hat{C}_{\alpha,T}(x_i) = [\hat{q}_{\alpha_{lo}}(x_i) - \omega_{lo_i}, \hat{q}_{\alpha_{hi}}(x_i) + \omega_{hi_i}]$\;
    \If{$i-T = 0$ mod $s$}{
    \For{$j=i-s,\ldots, i-1$}{
    Compute $E_{lo_j}$ and $E_{hi_j}$  as in Eq. (\ref{eq:assym_nc})\;
    $\mathcal{E}_{lo} = (\mathcal{E}_{lo} - \{E_{lo_{:s}}\}) \cup \{E_{lo_j}\}$ and reset index of $\mathcal{E}_{lo}$ \;
    $\mathcal{E}_{hi} = (\mathcal{E}_{hi} - \{E_{hi_{:s}}\}) \cup \{E_{hi_j}\}$ and reset index of $\mathcal{E}_{hi}$ \;}}}
    \BlankLine
    \Indm\Indmm
\Indp\Indpp
\caption{EnCQR}
\label{alg:enCQR}
\end{algorithm}
\DecMargin{1.2em}

\begin{figure}[!ht]
\centering
\includegraphics[width=.9\columnwidth]{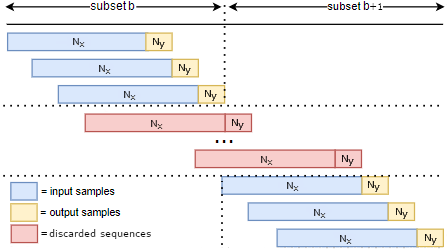}
\caption{Data split for training of ensemble learners.}
\label{fig:encqr-subsets} 
\end{figure}
Both EnbPI and EnCQR use homogeneous ensembles, but they differ in how training data subsets for the ensemble learners are built: EnbPI uses bootstrap samples drawn with replacement, while EnCQR creates $B$ disjoint subsets of length $T_b$, each used to train one learner (see Fig.~\ref{fig:encqr-subsets}). The trained learners can be used to produce out-of-sample residuals when applied to data that are not in the subset used for training. Within each subset, multiple overlapping input-output sequences (in blue-yellow) are extracted to be used as batches in the training of one learner. These have length $N = N_x + N_y \leq T_b$, where $N_x$ and $N_y$ denote the length of the input/output sequences. All sequences that go across subsets (in red) must be discarded for the residuals to remain valid. 
A large $N_x$ let learners capture long time dependencies in the data, while a large $B$ results in more learners and more robust predictions.
EnCQR benefits from large datasets (length $T$), as the number of residuals to perform CP is $T-B\cdot N_x$.
When $T$ is small, $N_x$ and $B$ should be reduced, trading the amount of residuals with the performance of the ensemble model.

Similarly to EnbPI, EnCQR utilizes an out-of-sample ensemble prediction strategy to apply CP to time series data, but replaces the aggregated point prediction of EnbPI with aggregated quantile predictions for the upper and lower bound of a PI.
As in CQR, the aggregated PI obtained using the ensemble learners are \textit{conformalized} using asymmetrical conformity scores
%, which can be seen as an extension of the conformity score used in the CQR algorithm. 
%The asymmetric conformity scores are 
defined as:
\begin{equation}
        E_{{{lo}}_i} = \hat{q}_{\alpha_{lo}}(x_i) - y_i\,, \quad
        E_{{hi}_i} = y_i - \hat{q}_{\alpha_{hi}}(x_i)\,.
    \label{eq:assym_nc}
\end{equation}
%Hence, while EnbPI is a ensemble version of CP that makes it applicable to time series, EnCQR is an equivalent extension of CQR.

The motivation for defining asymmetric conformity scores is that the distribution of the conformity scores for the two estimated conditional quantile functions can be skewed, resulting in the PI coverage error being asymmetrically spread over the left and right tails. 
If this occurs, the intervals can be wrongly conformalized, resulting in coverage below the target level. 
The asymmetric conformity score solves this problem by controlling the coverage of the two quantile functions independently. 
It can be shown that by using the asymmetric conformity score it is possible to obtain a stronger coverage guarantee compared to the original formulation, at the possible expense of slightly wider intervals~\cite{romano2019conformalized}.

Substitution of the point prediction in Eq.~\eqref{eq:enbpiPI} with the estimated quantile functions and the absolute conformity score with the asymmetric conformity score yields the EnCQR PI, which is defined as follows:
\begin{equation}
    \begin{split}
    \hat{C}_{\alpha,\phi}(x_t) = \left[\right. &\hat{q}_{\alpha_{lo}}(x_t) - Q_{1-\alpha}(\mathcal{E}_{lo})
    %(1-\alpha_{lo}) \text{ quantile of} \boldsymbol{E}_{lo}
    \,,\\ &\hat{q}_{\alpha_{hi}}(x_t) + Q_{1-\alpha}(\mathcal{E}_{hi})\left.\right]
    %(1-\alpha_{hi}) \text{ quantile of} \boldsymbol{E}_{hi} \big]
    \,,
    \label{eq:encqrPI}
    \end{split}
\end{equation}
where $\mathcal{E}_{lo} = \{E_{lo_i}\}_{i=t-1}^{t-T}$ and $\mathcal{E}_{hi} = \{E_{hi_i}\}_{i=t-1}^{t-T}$. 
As in CQR, the nominal quantile levels estimated by the underlying QR algorithm in EnCQR can be tuned to construct possibly PIs.

EnCQR combines the principles of EnbPI and CQR, while tackling the individual shortcomings of both methods. 
Specifically, EnCQR replaces the symmetric intervals in the EnbPI method with locally adaptive PIs by utilizing QR, as in CQR. 
Furthermore, EnCQR exploits the same moving window approach of EnbPI, by updating every $s$ predictions the conformity scores used to conformalize the QR intervals.
This allows to take into account new observations as they become available.
In addition, the update of the conformity scores and sequential construction of the PI makes EnCQR particularly suited for heteroscedastic time series.  

%%%%%%%%%%%%%%%%%%%%%%%%%%%%%%%%%%%%
%%% EXPERIMENTS
%%%%%%%%%%%%%%%%%%%%%%%%%%%%%%%%%%%%
\section{Experiments}

The experiments focus on comparing the PI obtained by EnCQR with methods based only on QR or CP, which use the same underlying regression algorithm.
To demonstrate the versatility of EnCQR we consider regression algorithms based both on deep learning architectures and traditional machine learning approaches.
Finding the best underlying regression algorithm for the specific dataset can improve the performance of both EnCQR and the competing methods, but not their \textit{relative difference} in performance. Therefore, finding such an optimal algorithm is outside the scope of our evaluation. The software implementation of EnQCR is available online\footnote{\url{https://github.com/FilippoMB/Ensemble-Conformalized-Quantile-Regression}}

\subsection{Datasets}
% We consider four real-world time series datasets from two application domains where heteroscedasticity is common. 
To test the effectiveness and generality of EnCQR in quantifying uncertainty in time series forecasting, we consider five real-world time series where the amount of heteroscedasticity and the seasonal patterns are significantly different.
We also considered the presence of exogenous variables as additional input time series, i.e., a multivariate input -- univariate output setting.
All the time series are partitioned into three disjoint sets, training, test, and validation, and are independently normalized by scaling the values to lie in the interval [0,1]. 
The time series from the first four datasets have hourly resolution and are reshaped into input-output pairs of size 168-24 (1week-1day) with a sliding window. In the last dataset, the input-output pairs have sizes 24-6.

The \textbf{Portugal} dataset\footnote{\url{https://archive.ics.uci.edu/ml/datasets/ElectricityLoadDiagrams20112014}} consists of time series of electricity consumption from 270 customers located in Portugal. 
The data have an hourly resolution from 2012 to 2014 and the consumption is measured in kiloWatt (kW). 
We arbitrarily selected 5 among the 270 time series. 
For each time series, the training/validation/test split is 12/12/12 months.

The \textbf{Elvia} dataset is shared by Elvia AS, a distribution system operator that operates grids in the Norwegian counties of Oslo, Viken and Innlandet. 
The dataset describes the electricity consumption for three classes of end-users; industry, household, and cabin. The data consists of two time series recorded hourly from 1 June 2018 to 1 June 2020 that contains the observed electricity load and the temperature forecast. 
One time series from each user category was arbitrarily selected. 
The training/validation/test split is 12/6/6 months (the 1\textsuperscript{st} year is used for training, while odd and even months of the 2\textsuperscript{nd} year are used as validation and test data, respectively). 

The \textbf{Solar} and \textbf{Wind} datasets\footnote{\url{https://github.com/Duvey314/austin-green-energy-predictor}} contain solar and wind power production data from Webberville Solar Farm and Hackberry wind farm in Texas, US. 
Data is recorded in MWh per hour from January 2017 to January 2020. 
Both datasets contain 6 ambient climate features. 
For both datasets, the training/validation/test split is 12/12/12 months. 

The \textbf{Temperature} dataset\footnote{\url{https://www.bgc-jena.mpg.de/wetter/}} contains weather data collected by the Max Planck Institute for Biogeochemistry. The dataset contains 15 meteorological variables, among which the ambient temperature, that is collected every 10 minutes from January 2009 to January 2016.
The dataset is subsampled to have an hourly resolution.
The first 80\% of the dataset is used for training, 10\% for validation, and 10\% for testing.  

The two power consumption and, in particular, the solar production and temperature time series are characterized by a strong seasonal pattern, which is not present in the wind power time series.
To measure the heteroscedasticity in each time series, we first calculate the standard deviation (std) of all values observed at a given hour of day over the entire dataset, and then the std of the resulting 24 values:
\[
\text{std}\left(\text{std}\{h_1d_1, h_1d_2, \ldots\}, \ldots, \text{std}\{h_{24}d_1, h_{24}d_2, \ldots\}\right),
\]
where, e.g., $h_1d_2$ is the measurement at hour 1 of day 2. 
A higher value indicates more variability between different hours, i.e.\ a higher degree of heteroscedasticity. Tab.~\ref{tab:datasetes_var} show a significant difference between the time series: the Portugal time series 250 has the lowest degree of variability, while the Solar and Temperature time series have the largest ones. 

% \begin{table}[!ht]
% \centering
% \small
% \begin{tabular}{c|c|c}
% \hline
% \multicolumn{2}{c}{Time series}  & \multicolumn{1}{c}{Variability}  \\
% \hline
% \multicolumn{1}{c|}{\multirow{5}{*}{\rotatebox{90}{Portugal}} }&{ID: 250}& $1.0e^{-4}$\\
% \cline{2-3}
% & ID: 77& $4.3e^{-4}$\\
% \cline{2-3}
% & ID: 50& $2.1e^{-4}$\\
% \cline{2-3}
% &ID: 90& $1.3e^{-4}$\\
% \cline{2-3}
% &ID: 27& $4.2e^{-4}$\\
% \hline
% \multicolumn{1}{c|}{\multirow{3}{*}{\rotatebox{90}{Elvia}} }
% & Industry& $9.3e^{-4}$\\
% \cline{2-3}
% & Household& $1.9e^{-4}$\\
% \cline{2-3}
% & Cabin& $1.9e^{-4}$\\
% \hline
% \multicolumn{2}{c|}{Solar}& $21.0e^{-4}$\\
% \hline
% \multicolumn{2}{c|}{Wind}& $3.0e^{-4}$\\
% \hline
% \end{tabular}
% \caption{Measure of heteroscedasticity in the time series of each dataset.}
% \label{tab:datasetes_var}
% \end{table}

\bgroup
\def\arraystretch{1} %vertical padding
\begin{table}[!ht]
\centering
\caption{Measure of heteroscedasticity for each dataset.}
\begin{tabular}{cc|c|cc|c}
\toprule
\multicolumn{2}{c}{Time series}  & \multicolumn{1}{c|}{Variability} & \multicolumn{2}{c}{Time series}  & \multicolumn{1}{c}{Variability}  \\
\hline
\multicolumn{1}{c}{\multirow{5}{*}{\rotatebox{90}{Portugal}} } & {ID: 250} & $1.0e^{-4}$ & \multicolumn{1}{c}{\multirow{3}{*}{\rotatebox{90}{Elvia}}    } & Industry& $9.3e^{-4}$\\
%\cline{2-3}
& ID: 77& $4.3e^{-4}$ & & Household& $1.9e^{-4}$\\
%\cline{2-3}
& ID: 50& $2.1e^{-4}$ & & Cabin& $1.9e^{-4}$\\
\cline{4-6}
&ID: 90& $1.3e^{-4}$ & \multicolumn{2}{c|}{Solar}& $21.0e^{-4}$\\
\cline{4-6}
&ID: 27& $4.2e^{-4}$ & \multicolumn{2}{c|}{Wind}& $3.0e^{-4}$\\
\cline{4-6}
& & & \multicolumn{2}{c|}{Temperature} & $39.1e^{-4}$\\
\bottomrule
\end{tabular}
\label{tab:datasetes_var}
\end{table}
\egroup

\subsection{Experimental Setup}
The experimental details are briefly summarized here. 
Further details are given in the Appendix. 

\bgroup
\def\arraystretch{1} %vertical padding
\begin{table*}[!ht]
\setlength\tabcolsep{.45em} %horizontal padding
\centering
\caption{PICP, PINAW, and CWC scores for EnCQR, QR, and EnbPI applied to different regression models. 
We report mean (and standard deviation) over 10 runs for NN-based and RF-based models and a single run for the deterministic SARIMA model.}
\begin{tabular}{c|c|c|ccc|ccc|ccc|ccc}
\toprule
\multicolumn{3}{c}{} & \multicolumn{3}{c}{\underline{LSTM}} & \multicolumn{3}{c}{\underline{TCN}} & \multicolumn{3}{c}{\underline{Random forest}} & \multicolumn{3}{c}{\underline{SARIMA}} \\
\multicolumn{1}{c}{\multirow{16}{*}{\rotatebox{90}{Portugal dataset}} } & \multicolumn{1}{c|}{} & Model & PICP & PINAW & CWC & PICP & PINAW & \multicolumn{1}{c|}{CWC}  & PICP & PINAW & CWC & PICP & PINAW & CWC \\
\hline
& \multirow{3}{*}{ID: 250}  
&  EnCQR  & .884 \t{(.011)} & .210 \t{(.011)} & \textbf{.783}
          & .890 \t{(.010)} & .210 \t{(.015)} & \textbf{.787}  
          & .900 \t{(.003)} & .219 \t{(.002)} & .781 
          & \multirow{3}{*}{.789} & \multirow{3}{*}{.154} & \multirow{3}{*}{.584} \\
& & QR    & .675 \t{(.030)} & .127 \t{(.032)} & .191
          & .718 \t{(.068)} & .165 \t{(.017)} & .309
          & .696 \t{(.016)} & .167 \t{(.013)} & .239 
          &  \\
& & EnbPI & .880 \t{(.005)} & .217 \t{(.012)} & .773 
          & .890 \t{(.013)} & .238 \t{(.011)} & .759
          & .876 \t{(.002)} & .202 \t{(.001)} & \textbf{.784}
          &  \\
\cline{2-15}
& \multirow{3}{*}{ID: 77}          
& EnCQR   & .900 \t{(.021)} & .249 \t{(.022)} & \textbf{.751}  
          & .900 \t{(.010)} & .197 \t{(.009)} & \textbf{.802}
          & .903 \t{(.003)} & .215 \t{(.002)} & \textbf{.784}
          & \multirow{3}{*}{.726} & \multirow{3}{*}{.179} & \multirow{3}{*}{.331} \\ 
& & QR    & .739 \t{(.063)} & .134 \t{(.028)} & .397  
          & .618 \t{(.044)} & .144 \t{(.013)} & .078
          & .589 \t{(.019)} & .124 \t{(.005)} & .048
          & \\
& & EnbPI & .901 \t{(.008)} & .297 \t{(.026)} & .702
          & .895 \t{(.017)} & .273 \t{(.020)} & .726 
          & .926 \t{(.002)} & .227 \t{(.002)} & .757
          & \\
\cline{2-15}
& \multirow{3}{*}{ID: 50}            
& EnCQR   & .890 \t{(.014)} & .252 \t{(.009)} & \textbf{.745} 
          & .900 \t{(.029)} & .326 \t{(.027)} & \textbf{.673} 
          & .911 \t{(.002)} & .285 \t{(.002)} & .712
          & \multirow{3}{*}{.802} & \multirow{3}{*}{.219} & \multirow{3}{*}{.585} \\
& & QR    & .838 \t{(.062)} & .242 \t{(.055)} & .675 
          & .758 \t{(.011)} & .211 \t{(.005)} & .430 
          & .766 \t{(.007)} & .193 \t{(.002)} & .470
          & \\
& & EnbPI & .905 \t{(.008)} & .299 \t{(.034)} & .700 
          & .897 \t{(.003)} & .344 \t{(.023)} & .655 
          & .913 \t{(.002)} & .275 \t{(.002)} & \textbf{.721}
          & \\
\cline{2-15}
& \multirow{3}{*}{ID: 90}       
& EnCQR   & .900 \t{(.021)} & .263 \t{(.016)} & \textbf{.737}   
          & .912 \t{(.023)} & .305 \t{(.025)} & .692
          & .914 \t{(.004)} & .295 \t{(.003)} & .700
          & \multirow{3}{*}{.893} & \multirow{3}{*}{.215} & \multirow{3}{*}{\textbf{.783}} \\
& & QR    & .610 \t{(.032)} & .153 \t{(.009)} & .067  
          & .816 \t{(.042)} & .232 \t{(.024)} & .621
          & .713 \t{(.015)} & .195 \t{(.005)} & .281
          & \\
& & EnbPI & .900 \t{(.005)} & .331 \t{(.011)} & .669
          & .926 \t{(.006)} & .288 \t{(.017)} & \textbf{.697}
          & .900 \t{(.002)} & .285 \t{(.002)} & \textbf{.715}
          & \\
\cline{2-15}
& \multirow{3}{*}{ID: 27}         
& EnCQR   & .910 \t{(.015)} & .225 \t{(.018)} & \textbf{.772}
          & .890 \t{(.012)} & .229 \t{(.015)} & \textbf{.768}
          & .911 \t{(.003)} & .189 \t{(.002)} & \textbf{.808}
          & \multirow{3}{*}{.590} & \multirow{3}{*}{.196} & \multirow{3}{*}{.044} \\
& & QR    & .800 \t{(.125)} & .184 \t{(.061)} & .604  
          & .898 \t{(.021)} & .240 \t{(.025)} & .759
          & .790 \t{(.007)} & .147 \t{(.002)} & .593
          & \\
& & EnbPI & .900 \t{(.002)} & .264 \t{(.014)} & .736
          & .895 \t{(.005)} & .277 \t{(.027)} & .722
          & .916 \t{(.002)} & .200 \t{(.002)} & .793
          & \\
\hline
\multirow{9}{*}{\rotatebox{90}{Elvia dataset}} 
& \multirow{3}{*}{Industry}             
& EnCQR   & .900 \t{(.010)} & .296 \t{(.023)} & .704
          & .900 \t{(.019)} & .303 \t{(.009)} & \textbf{.697}
          & .968 \t{(.003)} & .325 \t{(.005)} & .587
          & \multirow{3}{*}{.887} & \multirow{3}{*}{.430} & \multirow{3}{*}{.567} \\
& & QR    & .923 \t{(.010)} & .241 \t{(.017)} & \textbf{.747}
          & .927 \t{(.010)} & .293 \t{(.039)} & .691
          & .847 \t{(.017)} & .187 \t{(.005)} & \textbf{.747}
          & \\
& & EnbPI & .927 \t{(.005)} & .486 \t{(.021)} & .502 
          & .960 \t{(.006)} & .347 \t{(.048)} & .586
          & .979 \t{(.001)} & .339 \t{(.007)} & .548
          & \\
\cline{2-15}
&\multirow{3}{*}{Household}        
& EnCQR   & .909 \t{(.021)} & .242 \t{(.032)} & \textbf{.756}
          & .892 \t{(.035)} & .414 \t{(.041)} & .584
          & .949 \t{(.003)} & .457 \t{(.006)} & \textbf{.505} 
          & \multirow{3}{*}{.894} & \multirow{3}{*}{.148} & \multirow{3}{*}{\textbf{.851}} \\
& & QR    & .900 \t{(.023)} & .279 \t{(.072)} & .721 
          & .904 \t{(.015)} & .360 \t{(.032)} & \textbf{.639}
          & .764 \t{(.008)} & .157 \t{(.004)} & .484
          & \\
& & EnbPI & .980 \t{(.006)} & .369 \t{(.033)} & .520
          & .973 \t{(.004)} & .445 \t{(.018)} & .473
          & .998 \t{(.001)} & .513 \t{(.010)} & .365
          &  \\
\cline{2-15}
&\multirow{3}{*}{Cabin} 
& EnCQR   & .910 \t{(.022)} & .259 \t{(.027)} & \textbf{.738}
          & .906 \t{(.028)} & .401 \t{(.030)} & \textbf{.598}
          & .933 \t{(.004)} & .371 \t{(.004)} & \textbf{.609}
          & \multirow{3}{*}{.930} & \multirow{3}{*}{.257} & \multirow{3}{*}{.723} \\
& & QR    & .903 \t{(.057)} & .275 \t{(.072)} & .724
          & .861 \t{(.055)} & .447 \t{(.060)} & .528
          & .822 \t{(.009)} & .204 \t{(.006)} & .608
          & \\
& & EnbPI & .943 \t{(.008)} & .303 \t{(.016)} & .659
          & .960 \t{(.005)} & .574 \t{(.024)} & .382
          & .972 \t{(.002)} & .391 \t{(.004)} & .521
          & \\
\hline
\multicolumn{2}{c|}{\multirow{3}{*}{Solar}}
& EnCQR   & .900 \t{(.007)} & .354 \t{(.011)} & \textbf{.646}
          & .900 \t{(.009)} & .459 \t{(.028)} & .540
          & .917 \t{(.002)} & .310 \t{(.005)} & \textbf{.684}
          & \multirow{3}{*}{.842} & \multirow{3}{*}{.543} & \multirow{3}{*}{.413} \\
\multicolumn{1}{c}{}    
& & QR    & .910 \t{(.011)} & .365 \t{(.018)} & .633
          & .840 \t{(.046)} & .334 \t{(.029)} & \textbf{.597}
          & .863 \t{(.004)} & .298 \t{(.005)} & .673
          & \\
\multicolumn{1}{c}{}   
& & EnbPI & .910 \t{(.003)} & .631 \t{(.023)} & .367
          & .923 \t{(.006)} & .789 \t{(.019)} & .207
          & .906 \t{(.001)} & .622 \t{(.005)} & .377
          & \\
\hline
\multicolumn{2}{c|}{\multirow{3}{*}{Wind }}
& EnCQR   & .915 \t{(.009)} & .859 \t{(.025)} & .140 
          & .913 \t{(.004)} & .897 \t{(.015)} & .102
          & .906 \t{(.002)} & .799 \t{(.003)} & \textbf{.200}
          & \multirow{3}{*}{.980} & \multirow{3}{*}{1.00} & \multirow{3}{*}{.000} \\
\multicolumn{1}{c}{}    
& & QR    & .894 \t{(.004)} & .713 \t{(.012)} & .\textbf{286}
          & .775 \t{(.035)} & .682 \t{(.051)} & \textbf{.198}
          & .744 \t{(.008)} & .598 \t{(.006)} & .193
          &  \\
\multicolumn{1}{c}{}   
& & EnbPI & .901 \t{(.002)} & .974 \t{(.005)} & .025 
          & .913 \t{(.007)} & 1.00 \t{(.033)} & .000
          & .900 \t{(.002)} & .895 \t{(.006)} & .104
          & \\
          
\hline
\multicolumn{2}{c|}{\multirow{3}{*}{Temperature}}
& EnCQR   & .942 \t{(.010)} & .268 \t{(.018)} & \textbf{.692} 
          & .944 \t{(.009)} & .279 \t{(.055)} & \textbf{.680}
          & .943 \t{(.019)} & .337 \t{(.078)} & .621
          & \multirow{3}{*}{.813} & \multirow{3}{*}{.395} & \multirow{3}{*}{.482} \\
\multicolumn{1}{c}{}    
& & QR    & .820 \t{(.031)} & .198 \t{(.019)} & .611
          & .684 \t{(.105)} & .170 \t{(.031)} & .270
          & .931 \t{(.060)} & .365 \t{(.081)} & .555
          &  \\
\multicolumn{1}{c}{}   
& & EnbPI & .940 \t{(.008)} & .341 \t{(.022)} & .627 
          & .932 \t{(.004)} & .335 \t{(.056)} & .645
          & .941 \t{(.012)} & .305 \t{(.073)} & \textbf{.658}
          & \\
\bottomrule
\end{tabular}
\label{tab:result}
\end{table*}
\egroup

\begin{table}[!ht]
\centering
\caption{Average CWC across all datasets and regression algorithms.}
\begin{tabular}{ccccc}
\toprule
                              & EnCQR & QR   & EnBPI & SARIMA \\ \hline
\multicolumn{1}{c|}{Avg. CWC} & \textbf{.648}  & .477 & .558  & .486 \\ 
\bottomrule
\end{tabular}
\label{tab:result_avg}
\end{table}

\paragraph{Reference models} 
We compare EnCQR with EnbPI and QR.
All methods are based on the same underlying regression algorithms. 
Additionally, we include the SARIMA model as a representative of traditional statistical models for time series forecasting.
SARIMA is meant to be a baseline rather than a competitor of EnCQR. As it cannot model heteroscedasticity, it produces PIs that are not adaptive and are valid only under restrictive assumptions.

All models are designed to construct multi-step probabilistic forecasts in the form of 90\% PIs, i.e.\ $\alpha = 0.10$. 
% Specifically, PIs for the next 24 hours of each day in the test sets are constructed using the observations from the previous 168 hours.

\paragraph{Regression algorithms} 
We perform experiments with quantile random forest (QRF) and QRNNs as the underlying regression algorithms. 
Given the prominence of NNs in recent time series forecasting research, we decided to use two different QRNN architectures: Temporal Convolutional Network (TCN)~\cite{chen2020probabilistic}, and Long Short-term Memory Neural Network (LSTM)~\cite{wang2019probabilistic}.

\paragraph{Model training}
We conduct a random search over the NN hyperparameters, detailed in the Appendix. 
All NNs are trained to minimize the pinball loss. 
The ensemble NNs in EnbPI are trained to predict only the 0.5 quantile. 
The QRNN and EnCQR models are trained to predict, at the same time, multiple quantile levels, 0.95, 0.50 and 0.05, rather than fitting individual network instances for each quantile. 
This is done by modifying the pinball loss in~\eqref{eq:pinball}, which is averaged over all observations and all target quantiles. 
Early stopping terminates the training process if the pinball loss on the validation set does not improve over 50 consecutive epochs. For the QRF models, we build 10 trees that are expanded until all leaves are pure or until all leaves contain less than 2 samples, and we estimate the 0.95, 0.50 and 0.05 quantiles.

\paragraph{Ensemble learners}
Based on the size of the datasets, we chose $B=3$ to be a suitable number of ensemble learners in EnbPI and EnCQR.
While using more learners could give better performance, for a fixed amount of data each of the $B$ subsets would become smaller (see Fig.~\ref{fig:encqr-subsets}). As a consequence, the learners would be trained on less data and might not capture longer temporal dependencies.

The window size parameter $s$ should reflect the nature of the data.
Since all time series have hourly resolution and a strong daily seasonality, we set $s = 24$.
Finally, all ensemble models use the mean as aggregation function $\phi$.
%TCNs and LSTM networks require the input samples to be presented sequentially to learn the temporal structures within the time series, and sampling with replacement is not suited. Therefore, the ensemble subsets for all regression algorithms are created as in lines 1-4 of Alg.~\ref{alg:enCQR}. 

\paragraph{Evaluation metrics} 
To evaluate the quality of the PIs, both their coverage and width must be quantified. For this we use two measures; the \textit{prediction interval coverage probability} (PICP)~\cite{shepero2018residential_picp}: 
\begin{equation*}
    \text{PICP} = \frac{1}{n}\sum_{i=1}^{n}c_i, \quad c_i =     
    \begin{cases}
        1, & y_i \in [L_i,U_i]\\
        0, & y_i \notin [L_i,U_i]
\end{cases}
\end{equation*}
and \textit{prediction interval normalized average width} (PINAW):
\begin{equation*}
    \text{PINAW} = \frac{1}{n R}\sum_{i=1}^{n}(U_i - L_i), \quad  R = y_\text{max} - y_\text{min} \,.
\end{equation*}
Here $U$ and $L$ denote the upper and lower bound of the PI. 
PICP alone is not sufficient to measure performance, since very wide PIs have high coverage but are less informative. 
An optimal PI has a PICP close to the designed confidence level and minimizes, at the same time, the PINAW.
To summarize with a single value the quality of the PI, we adopt a modification of the \textit{coverage width-based criterion} (CWC)~\cite{shen2018wind} that penalizes under- and overcoverage in the same way:
\begin{equation}
    \label{eq:cwc}
    \text{CWC} = (1-\text{PINAW})e^{ -\eta(\text{PICP} - (1-\alpha))^2},
\end{equation}
where $\eta$ is a user-defined parameter that balances the PINAW and PICP contributions.
In our experiment, we set $\eta=30$.

\subsection{Results and Analysis}
Table \ref{tab:result} summarizes the results obtained on each dataset using the NN-based and RF-based models. 
We also report the results of SARIMA for comparison.
For each dataset and regression algorithm, we highlight in bold the best result in terms of CWC.
Table \ref{tab:result_avg} reports the mean CWC of the different approaches across all datasets and regression algorithms.

\subsubsection{PI Coverage and PI Width}
\begin{figure}[!ht]
\centering
\includegraphics[width=0.8\columnwidth]{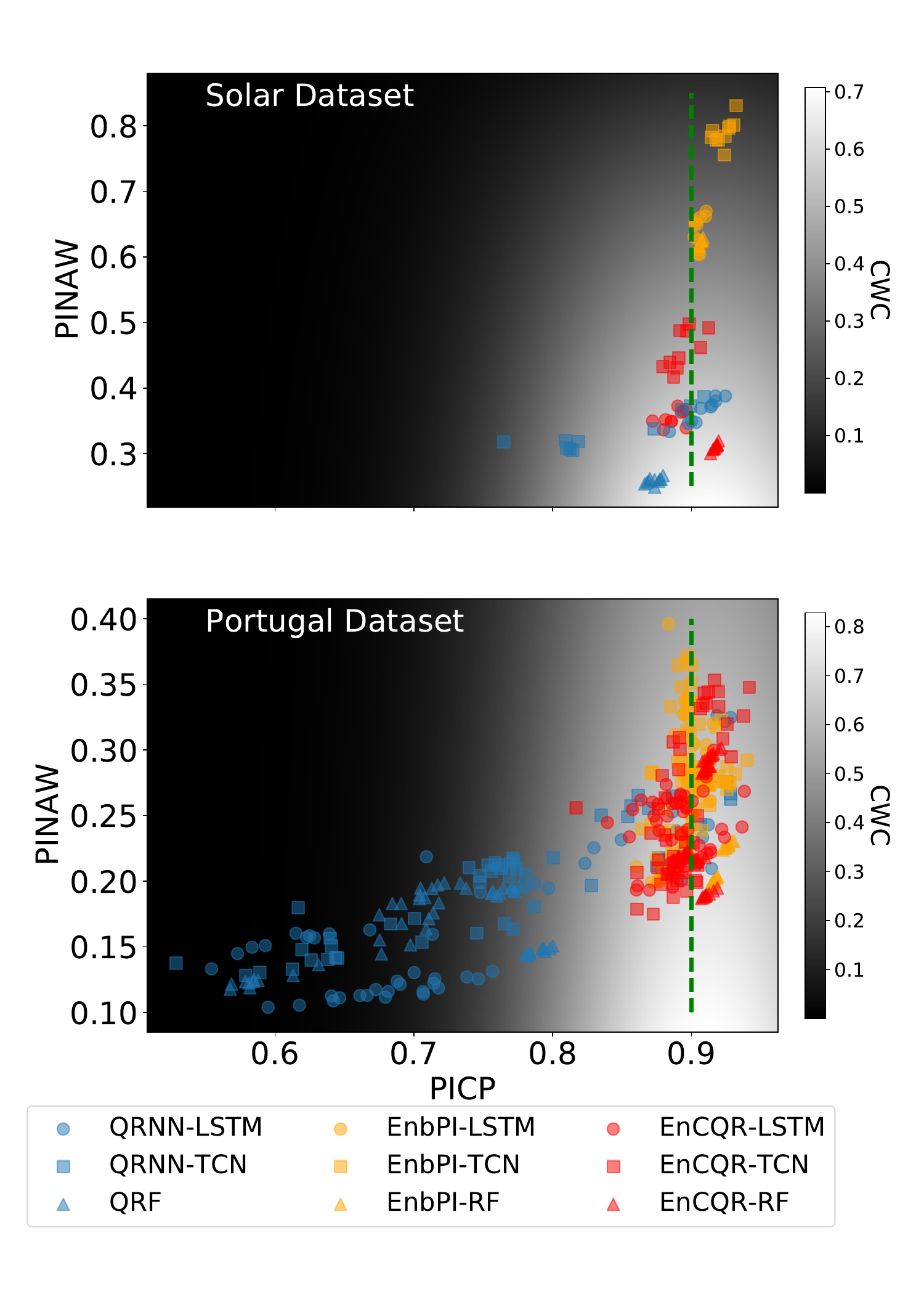}
\caption{PICP versus PINAW for all instances of the NN an RF models for the Solar and Portugal datasets. 
The background shows the value of CWC at each region: the brighter the better.}
\label{fig:scatter} 
\end{figure}

Both EnCQR and EnbPI successfully construct approximately valid PIs for all time series, demonstrating that they work well for different data distributions, regardless of the regression algorithm used in the ensemble.  
However, the width of the PIs constructed by EnbPI considerably varies compared to EnCQR, which produces the sharpest valid PIs.
This can be explained by referring to the variability measures presented in Table~\ref{tab:datasetes_var}; 
For time series with a low degree of heteroscedasticity, the quality of the PIs constructed by EnCQR and EnbPI is approximately equal, whereas for the more heteroscedastic time series, EnCQR constructs significantly sharper and more informative PIs.
This is illustrated in Fig.~\ref{fig:scatter}, which plots PICP against PINAW for the NN- and RF-based models on the Solar and Portugal datasets.
For both datasets, nearly all EnCQR instances in Fig.\ \ref{fig:scatter} are located in the bottom-right corner, indicating that these models produce PIs with the highest PICP and narrowest PINAW. 
For the Portugal dataset, EnbPI achieves a similar coverage level as EnCQR, but the PINAW in EnbPI is generally higher, indicating that the PIs are wider and less informative. 
For the Solar dataset, which is more heteroscedastic, the difference in PINAW between the EnbPI and EnCQR is even larger.

SARIMA, QRF, and QRNN yield PIs whose PICP greatly varies in different datasets, as they lack robustness and do not enjoy the coverage guarantee of the CP-based models.
In fact, they might construct very narrow intervals where the actual coverage of the PIs is significantly lower than the desired confidence level.

The results discussed so far are aligned with the CWC values in Table \ref{tab:result}.
Table \ref{tab:result_avg} shows that when CWC is averaged across all dataset and regression algorithms EnCQR achieves the top performance and EnBPI, which comes second, achieves a significantly lower average CWC score. 
In the next analyses we focus only on EnCQR and EnbPI, since SARIMA, QRF and QRNN often fail to produce a valid coverage.

\subsubsection{Symmetric vs. Adaptive PIs}

\begin{figure*}[!hbtp]
    \centering
    \includegraphics[width = 0.7\textwidth]{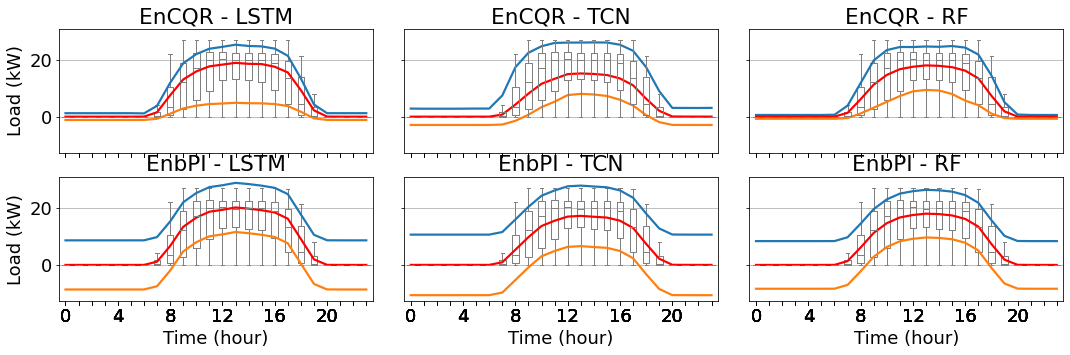}
    \caption{Average PIs for all hours in the solar test dataset. The red line represent the predicted mean, and the blue and yellow line represent the average upper and lower PI bounds, respectively. All lines show the aggregated results from ten individual runs. The underlying boxplot represent the hourly variation for the observations in test datasets.}
    \label{fig:industry_pi_box}
\end{figure*}
Here, we study how producing symmetric PIs rather than locally adaptive PIs affects the quality of the results.
The analysis is performed using the \textit{Solar} time series, which has a strong seasonal pattern since most of the energy is produced at the central hours of the day. 
The variation in hourly energy production in Fig.\ \ref{fig:industry_pi_box}, clearly indicates the presence of heteroscedasticity in the time series.

As previously stated, both EnCQR and EnbPI use a sliding window of size $s=24$ to reflect the data collection process. 
As a result, the set of leave-one-out residuals is updated after every 24 hours.
Consequentially, the width of the PIs generated by EnbPI is fixed for the 24-hours interval, indicating an equal amount of uncertainty for all the hours in a day.
On the other hand, the adaptive EnCQR intervals show that there is significantly lower uncertainty during the night. 
The advantage of locally adaptive intervals for heteroscedastic data is clear: EnCQR constructs significantly more informative intervals for the hours with less variability. 

In EnbPI to guarantee valid coverage with symmetric intervals when the data is heteroscedastic, the length of the intervals must increase significantly  to include points far from the mean value. 
The length increases identically in both directions, which is undesirable if the spread from the mean is not symmetric. 
This is the case of the Solar time series: in the middle of the day, the deviation from the mean is greater in the downwards direction due to the potential absence of sun. 
On the other hand, the PI of EnCQR are not constrained to be symmetric and can, therefore, capture in which direction the variability is larger. 
Referring to Fig.\ \ref{fig:industry_pi_box}, EnCQR-LSTM correctly captures the variability in the time series: the predicted 0.50 quantile is almost perfectly aligned with the median of the boxes, while the lower PI bound extends downwards much further than the upper PI bound.

\subsubsection{The Effect of Conformalization}
\begin{figure}[!ht]
    \centering
    \includegraphics[width=0.7\columnwidth]{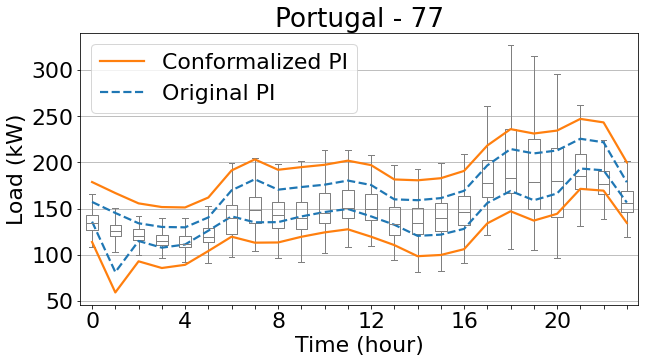}
    \caption{Average original and conformalized PIs for all hours in the Portugal - 77 test set. The PICP of the original PI is  improved by the conformalization. }
    \label{fig:conf_int_port}
\end{figure}

In EnCQR, the PI constructed by the ensemble of QR learners is conformalized by adding or subtracting an error term to the interval's width.
This error term quantifies the accuracy of the original interval and addresses both under- and overcoverage, since the PIs can be extended or shortened to improve both PI coverage and width. 
To analyze the effect of conformalization, we use the results of the EnCQR-LSTM model for Station 77 from the Portugal dataset, the Industry user in the Elvia dataset, and the Solar dataset. 
Fig.~\ref{fig:conf_int_port} depicts the conformalized and original intervals of the time series associated with station 77 and clearly shows the improvement of conformalization. 
The original PIs do not cover the boxes of the underlying boxplot, which extends from the 1\textsuperscript{st} to the 3\textsuperscript{rd} quartile, hence indicating a significant undercoverage. 
The conformalization extends both the upper and lower PI bound and guarantees a valid coverage. 
The results for the other datasets are summarized in Table~\ref{tab:conformalization_int}.

\bgroup
\def\arraystretch{1} %vertical padding
\begin{table}[!ht]
\centering
\caption{PICP and PINAW for the original and conformalized PIs for the EnCQR-LSTM models.}
\begin{tabular}{lcc}
\toprule
\multicolumn{1}{c}{}&\multicolumn{2}{c}{\underline{PICP / PINAW}}
\\
\multicolumn{1}{c}{Time series}&\multicolumn{1}{c}{conformalized PI}&\multicolumn{1}{c}{original PI}  \\
\hline
{Portugal - 77}&   0.900 / 0.249 & 0.524 / 0.109 \\
{Elvia - Industry} & 0.900 / 0.296 & 0.550 / 0.157   \\
{Solar}& 0.900 / 0.354 & 0.724 / 0.272 \\ 
\bottomrule
\end{tabular}
\label{tab:conformalization_int}
\end{table}
\egroup

%%%%%%%%%%%%%%%%%%%%%%%%%%%%%%%%%%%%
%%% CONCLUSIONS
%%%%%%%%%%%%%%%%%%%%%%%%%%%%%%%%%%%%
\section{Discussion and conclusions}

QR-based models tend to produce overly confident and often invalid PIs, since they are too narrow and the actual observations fall, on average, outside the PI boundaries more often than the specified confidence level. 
This lack of robustness  motivates the need of probabilistic frameworks, such as CP, to obtain valid PIs. 
The recently proposed EnbPI method allows to apply CP to time series data. 
However, despite of the advantage of valid coverage, CP tends to be unnecessarily conservative as it constructs PIs with constant length, which are uninformative especially when dealing with heteroscedastic data. 

In this paper we proposed EnCQR, a probabilistic time series forecasting method that leverages CP to generate PIs with valid coverage and ensemble learners performing QR to handle heteroscedastic data. 
Experiments on real-world datasets with different degrees of heteroscedasticity demonstrated the superior performance of the proposed method compared to methods based only on CP or QR. 
Our method outperforms CP-based models such as EnbPI in terms of PI sharpness and QR-based models in terms of PI coverage.
For homoscedastic data, EnCQR performs approximately equal to EnbPI in terms of PI quality.
For heteroscedastic data, EnCQR outperforms EnbPI as the PI width adapt well to local variability.

The most appealing property of EnCQR is unarguably that the PIs are guaranteed to marginally satisfy the designed coverage rate for finite samples and are adaptive to local variability, hence construing sharper PIs compared to other CP-based methods. 
EnCQR is particularly suitable for large datasets, due to the need to create independent subsets of consecutive data. 
To have enough ensemble learners, while ensuring that they are trained on enough data and capture long-range dependencies, longer time series are warranted.

An advantage of EnCQR is that it can be applied on top of any ensemble of QR models, such as the two ``standard'' neural network architectures (LSTM and TCN) trained with the pinball loss. % that we used in our experiments.
It is worth mentioning that several advanced deep learning models for probabilistic forecasting have been proposed in the past few years~\cite{MASHLAKOV2021116405} and some of them are readily available in open-source libraries~\cite{alexandrov2020gluonts}.
Despite producing accurate forecasts, none of these methods can generate PIs that are both valid and adaptive.
Therefore, an interesting future work would be to replace the LSTM and TCN backbones with more powerful models to obtain narrower, yet valid and adaptive, PIs.

As a final remark, this paper focused on frequentist approaches to compute PIs, but uncertainty and interval quantification can also be estimated using Bayesian approaches.
Bayesian  uncertainty  estimates  often  fail  to capture  the  true  data  distribution, due to model bias that does not assign the right probability to every credible interval~\cite{10.5555/3295222.3295387}.
On the other hand, Bayesian methods are more robust when the data available are scarce and the model is very uncertain about the prediction.
A principled and fair comparison between Bayesian and the proposed frequentist intervals must be done with caution and it might be interesting to explore in future work.

%%%%%%%%%%%%%%%%%%%%%%%%%%%%%%%%%%%%
%%% APPENDIX 
%%%%%%%%%%%%%%%%%%%%%%%%%%%%%%%%%%%%
% \clearpage
\appendix

\subsection{Neural Networks implementation details}
The neural network-based models are implemented in Tensorflow \cite{abadi2016tensorflow,chollet2015keras}.
The optimal network hyperparameters, such as learning rate, batch size, and layer units, are identified by performing a random hyperparameter search.
More specifically, we randomly select different parameter configurations from specified intervals and we select the configuration that achieves the highest performance (in terms of quality of the prediction interval) on the validation dataset.

All neural networks are trained using the Adam optimizer~\cite{kingma2014adam}. 
L2 regularization is used to prevent overfitting and improve the generalization capabilities of the network. 
The same L2 penalty regularization is applied to the input, hidden and output weights. 
For all networks, the $\lambda_2$ value (which specifies the contribution to the loss of the term that penalizes the L2 norm of the weights) is optimized during the hyperparameters search.
In particular, the value of $\lambda_2$ is randomly sampled from the interval $[0,0.1]$ using a logarithmic scale. 
The configurations of the specific deep learning models are described in the following.

\subsubsection{TCN} 
\begin{figure}[!ht]
\centering
\includegraphics[width = .8\columnwidth]{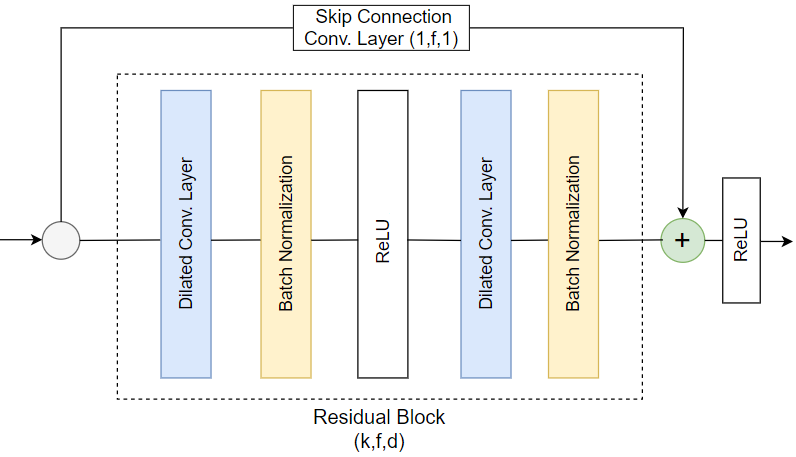}
\caption[Temporal convolutional neural network residual block]{TCN residual block, containing two identical dilated convolutional layers with kernel size = k, no. convolutional filters = f, and dilation factor = d. The skip connection consists of a convolutional layer with k = 1, d = 1, and the same number of filter as the convolutional layers within the residual block, followed by an element-wise addition $\oplus$}
\label{fig:tcn_resblock}
\end{figure}

The TCN setup, inspired by the DeepTCN\footnote{https://github.com/oneday88/deepTCN} network presented by \cite{chen2020probabilistic}, consists of several stacked residual blocks containing dilated convolutional layers, followed by a final fully-connected layer that maps the output of the residual blocks into quantile predictions. 
The residual blocks, illustrated in Fig.\ \ref{fig:tcn_resblock}, consist of two identical dilated causal convolutional layers, both followed by a batch normalization layer and ReLU activation. 
Depending on the number of residual blocks in the network, the residual block's output is either passed as input to the next residual block or to the final fully-connected output layer. 
Contrarily to the skip connections in ordinary residual networks, the skip connections in the TCN residual blocks contain a $1\times 1$ convolutional layer with the same number of filters of the convolutional layers in the residual block, prior to the element-wise sum operation $\oplus$. 
The additional convolutional layer ensures that the sum operator receives tensors of the same shape, as the input and output of the TCN residual block can have different widths \cite{bai_empirical}.

The optimal values of the TCN hyperparameters are searched within the following ranges: 
\begin{itemize}
    \item Dilation factor $d = 2^i, i \in [0,5]$
    \item Kernel size $k = [2,7]$
    \item No. filters $N_f = [2,200]$
    \item Learning rate $\eta = [0.0001,0.01]$
    \item L2 regularization parameter $\lambda_2 = [0.0001,0.1]$
    \item Batch size $B = [16,32,56,64]$
    \item Quantile ranges from the EnCQR models: \newline
    $\hat{q_{lo}} = [0.01,0.2], \hat{q_{hi}} = [0.7,0.99]$
\end{itemize} 
The optimal hyperparameter values found are reported in Tab.\ref{tab:HPNN}

\subsubsection{LSTM}
The LSTM network contains one or more hidden LSTM layers, followed by a fully-connected output layer used to map the output from the LSTM layers into the actual quantile forecasts. 
All hidden layers have the same number of units. 
When stacking several LSTM layers, the output from the first LSTM layer is used as input to the next. 
%Stacking several LSTM layers makes the network deeper, and the field of deep learning is built around the idea of deeper network being more effective at representing some functions compared to shallower networks \cite{bengio2009learning}.
Stacking several recurrent levels allows each recurrent layer to operate at different timescales \cite{pascanu2013deepRNN}, which often improves the network performance in sequence prediction problems.

Tensorflow provides both stateful and stateless LSTM cells\footnote{\url{https://keras.io/api/layers/recurrent\_layers/lstm/}}. 
In a stateful LSTM, the last state for each sample at index $i$ in a batch will be used as initial state for the sample of index $i$ in the following batch. In a stateless LSTM layer, the hidden states are reset after each batch and, therefore, the network cannot learn time dependencies spanning across different batches. The LSTM networks implemented in the experiments are stateful and the batch size is, therefore, fixed for all LSTM networks. 

The optimal values of the LSTM hyperparameters are searched within the following ranges: 
\begin{itemize}
    \item No. units in hidden layers $N_u = [8,200]$
    \item No. hidden layers $N_h = [1,3]$
    \item Learning rate $\eta = [0.0001,0.01]$
    \item L2 regularization parameter $\lambda_2 = [0.0001,0.1]$
    \item Quantile ranges from the EnCQR models: \newline
    $\hat{q_{lo}} = [0.01,0.2], \hat{q_{hi}} = [0.7,0.99]$
\end{itemize} 
The optimal hyperparameter values are reported in Tab.\ref{tab:HPNN}

\bgroup
\def\arraystretch{.9} %vertical padding
\begin{table*}[!ht]
\setlength\tabcolsep{.45em} %horizontal padding
\centering
\caption{Optimal model configurations for NN-based models for all datasets. The acronyms in the table are: $N_u$: number of units in LSTM layers, $N_h:$ number of hidden layers, $\eta$: learning rate, $\lambda_2$: L2 regularization parameter, b: batch size, $d$: dilation factor, $N_f$: number of filters in convolutional layers in each residual block, $k$: size of convolutional kernel, and $\hat{q}_{lo}$ and $\hat{q}_{hi}$ are the lower and upper quantile levels predicted by the underlying NN in EnCQR.}
\begin{tabular}{c|c|c|ccccc|ccccccc}
\toprule
\multicolumn{3}{c}{} & \multicolumn{5}{c}{\underline{LSTM}} & \multicolumn{7}{c}{\underline{TCN}} \\
\multicolumn{1}{c}{\multirow{16}{*}{\rotatebox{90}{Portugal dataset}} } & \multicolumn{1}{c}{TS} & \multicolumn{1}{|c}{Model}   & $N_u$& $N_h$  & $\eta$ & $\lambda_2$& $\hat{q}_{lo}$, $\hat{q}_{hi}$& d & $N_f$ & k&$\eta$ & $\lambda_2$&B& $\hat{q}_{lo}$, $\hat{q}_{hi}$ \\
\hline
&\multirow{3}{*}{ID: 250}  
                 & EnCQR &  5 & 2 & 1.0$e^{-3}$ & 5.0$e^{-3}$ & [0.05,0.92]& 0 & 6 & 7 & 7.0$e^{-3}$&5.0$e^{-3}$&32&[0.20, 0.78]\\
                 & &QRNN  &   57 & 2 & 1.8$e^{-3}$& 2.5$e^{-3}$ & - &  3 & 18&6&0.025&5.0$e^{-3}$&64&- \\
                 &  & EnbPI & 68 & 3 & 1.0$e^{-3}$&5.0$e^{-3}$ &- & 1& 5 & 7 & 5.0$e^{-3}$ &5.0$e^{-3}$&32& -     \\
\cline{2-15}
&\multirow{3}{*}{ID: 77}          
                 & EnCQR & 48 & 1 & 1.4$e^{-4}$ & 5.0$e^{-3}$ & [0.05,0.95]&0 & 6 & 7 & 7.0$e^{-3}$&5.0$e^{-3}$&32&[0.20, 0.78]\\
                 & &QRNN  & 49 & 2 & 2.5$e^{-4}$ & 1.8$e^{-4}$ & - & 3 & 13 & 6 & 2.6$e^{-3}$ & 0.01 & 32 & -     \\
                 &  & EnbPI & 59 & 3 & 7.9$e^{-4}$ & 5.0$e^{-3}$ & - & 1 & 8 & 7 & 5.0$e^{-3}$ & 5.0$e^{-3}$&32&-    \\
\cline{2-15}
&\multirow{3}{*}{ID: 50}            
                 & EnCQR & 89 & 1 & 1.0$e^{-3}$ & 5.0$e^{-3}$ & [0.09,0.89] & 0 & 50 & 7 & 7.0$e^{-4}$ & 5.0$e^{-3}$ & 32 & [0.19, 0.87]\\
                 & &QRNN  & 93 & 2 & 7.5$e^{-4}$ & 4.6$e^{-4}$& - & 4 & 8 & 7 & 7.0$e^{-4}$ & 1.6$e^{-3}$& 32 & -      \\
                 &  & EnbPI & 66 & 2 & 2.0$e^{-3}$&5.0$e^{-3}$&-& 1 & 2  & 7 & 2.0$e^{-3}$ & 0.05 & 32 & -    \\
\cline{2-15}
&\multirow{3}{*}{ID: 90}       
                 & EnCQR & 69& 2& 1.2$e^{-3}$&1.0$e^{-3}$&[0.06,0.91]& 1 & 50 & 7 & 1.0$e^{-3}$ & 5.0$e^{-3}$ & 32 & [0.08, 0.92] \\
                 & &QRNN  &  49 & 2 & 2.5$e^{-4}$ & 1.8$e^{-4}$ & - & 2 & 8 & 7 & 7.0$e^{-4}$ & 1.6$e^{-3}$ & 32 & -   \\
                 &  & EnbPI &  99 & 2 & 3.5$e^{-3}$&5.0$e^{-3}$&-&  1 & 9 & 7 & 9.0$e^{-3}$ & 0.05 & 32 & - \\
\cline{2-15}
&\multirow{3}{*}{ID: 27}         
                 & EnCQR &  48 & 1 & 1.4$e^{-3}$&5.0$e^{-3}$ & [0.14,0.94] & 0 & 50 & 7 & 1.0$e^{-3}$ & 5.0$e^{-3}$& 32 & [0.1, 0.97]\\
                 & &QRNN  &    166 & 3 & 2.4$e^{-4}$ & 0.058 & - &  0 & 23 & 7 & 1.0$e^{-3}$ & 1.0$e^{-3}$ & 32 & - \\
                 &  & EnbPI &  47 & 3 & 1.0$e^{-4}$ & 5.0$e^{-3}$ & - & 1 & 7 & 7 & 8.0$e^{-3}$&0.05 & 32 & -    \\
\hline
\multirow{9}{*}{\rotatebox{90}{Elvia dataset}} 
&\multirow{3}{*}{Industry}             
                 & EnCQR & 10 & 1 &1.0$e^{-3}$ & 0.01 & [0.10,0.80] & 0 & 3 & 7 & 2.5$e^{-3}$ & 5.0$e^{-3}$& 32&[0.02, 0.76] \\
                 & &QRNN  &  100 & 2 & 1.0$e^{-4}$ & 1.0$e^{-3}$ & - & 2 & 2 & 5 & 0.01 & 0.05 & 32 & -     \\
                 &  & EnbPI & 27 & 1 & 5.0$e^{-3}$ & 5.0$e^{-3}$ & - & 0 & 15 & 7 & 0.01& 5.0$e^{-3}$& 32 & -    \\
\cline{2-15}
&\multirow{3}{*}{Household}        
                 & EnCQR & 16 & 1 & 0.011 &5.0$e^{-3}$& [0.17,0.79] & 0 & 3 & 7 & 0.01 & 0.01 & 32 & [0.10, 0.91] \\
                 & &QRNN  &   122 & 2 & 1.0$e^{-4}$ & 8.6$e^{-4}$ & - & 0 & 3 & 4 & 0.01 & 2.3$e^{-3}$ & 56 & -   \\
                 &  & EnbPI &   27 & 1 & 5.0$e^{-3}$ & 5.0$e^{-3}$ & - & 0 & 49 & 7 & 0.01 & 5.0$e^{-3}$ & 32 & -   \\
\cline{2-15}
&\multirow{3}{*}{Cabin} 
                 & EnCQR & 17 & 1 & 0.024 & 5.0$e^{-3}$&[0.12,0.85] & 0 & 90 & 3 & 6.0$e^{-3}$ & 1.0$e^{-5}$ & 32 & [0.04, 0.93] \\
                 & &QRNN  & 138 & 3 & 9.8$e^{-4}$ & 0.029 & - & 1 & 2 & 5 & 2.3$e^{-3}$ & 1.8$e^{-4}$ & 64 &  -    \\
                 &  & EnbPI &  59 & 1 & 9.0$e^{-3}$ & 5.0$e^{-3}$ & - & 0 & 200 & 7 & 9.0$e^{-3}$ & 5.0$e^{-3}$& 32 & -  \\
\hline
\multicolumn{2}{c|}{\multirow{3}{*}{Solar}}
                 & EnCQR & 89 & 1 & 9.0$e^{-4}$ & 5.0$e^{-3}$ & [0.09,0.89] & 1 & 101 & 7 & 1.8$e^{-3}$ & 5.0$e^{-3}$ & 32 &[0.15, 0.99]   \\
\multicolumn{1}{c}{} & &QRNN  &  18 & 1 & 5.0$e^{-3}$ & 5.0$e^{-3}$& - & 2 & 5 & 7 & 3.5$e^{-3}$ & 5.0$e^{-3}$ & 32 & -   \\
\multicolumn{1}{c}{} &  & EnbPI &  147 & 2 & 1.0$e^{-3}$ & 5.0$e^{-3}$ & - & 2 & 5 & 7 & 3.5$e^{-3}$ & 5.0$e^{-3}$ & 32 & -    \\
\hline 
\multicolumn{2}{c|}{\multirow{3}{*}{Wind}}
                 & EnCQR & 69 & 1 & 9.0$e^{-4}$ & 5.0$e^{-3}$ & [0.05,0.99] & 0 & 79 & 7 & 5.0$e^{-3}$ & 5.0$e^{-3}$ & 32 & [0.05, 0.92]  \\
\multicolumn{1}{c}{} & &QRNN  &  18 & 3 & 5.0$e^{-3}$ & 1.0$e^{-3}$ & - & 1 & 12 & 7 & 2.5$e^{-3}$ & 5.0$e^{-3}$ & 32 & -    \\
\multicolumn{1}{c}{} &  & EnbPI & 47  &3 &1.0$e^{-4}$& 5.0$e^{-3}$& - & 1 & 12 & 7 & 2.5$e^{-3}$ & 5.0$e^{-3}$ & 32 & -     \\
\hline 
\multicolumn{2}{c|}{\multirow{3}{*}{Temperature}}
                       & EnCQR  & 32 & 1 & 5.0$e^{-4}$ & 1.0$e^{-5}$ & [0.15, 0.85] & 4 & 32 & 7 & 5.0$e^{-4}$ & 1.0$e^{-5}$ & 32 & [0.10, 0.91]  \\
\multicolumn{1}{c}{} & & QRNN   & 32 & 1 & 5.0$e^{-4}$ & 1.0$e^{-5}$ & -            & 4 & 32 & 7 & 2.5$e^{-3}$ & 1.0$e^{-4}$ & 32 & -    \\
\multicolumn{1}{c}{} & & EnbPI  & 32 & 2 & 5.0$e^{-4}$ & 1.0$e^{-4}$ & -            & 4 & 64 & 7 & 5.0$e^{-4}$ & 5.0$e^{-4}$ & 32 & -     \\
\bottomrule
\end{tabular}
\label{tab:HPNN}
\end{table*}
\egroup
    
\subsection{SARIMA implementation details}
The seasonal ARIMA models are implemented using the SARIMAX function from the \textit{statsmodels}\footnote{\url{www.statsmodels.org}} Python library. The SARIMAX function fits a model using the provided training data, and predicts a specified number of out-of-sample forecasts from the end of the training samples index. For the multivariate dataset, both historical load and historical records of the exogenous variables are presented to the model. The model orders for each time series are determined by analysing the ACF and PACF plots and by using the AIC criterion to select the optimal model. 
The model orders for all time series are reported in Table \ref{tab:model_par_SARIMA}. 

\begin{table}[ht!]
\centering
\caption{Seasonal ARIMA model parameters, (p,d,q)$\times$(P,D,Q)m. In Elvia, Solar, Wind, and Temperature datasets exogenous variables are added and the extended model is termed SARIMAX.}
\begin{tabular}{lc}
\toprule
\multicolumn{1}{c}{}&\multicolumn{1}{c}{\textbf{Portugal}}  \\
ID: 250        &         SARIMA (3,1,1)$\times$(1,1,1)24                \\

ID: 77         &         SARIMA (2,0,2)$\times$(1,1,1)24    \\
ID: 50         &           SARIMA (2,1,4)$\times$(1,1,1)24      \\  

ID: 90         &           SARIMA (4,0,0)$\times$(1,1,1)24            \\

ID: 27         &           SARIMA (3,1,2)$\times$(1,1,1)24            \\
\hline
\multicolumn{1}{c}{}&\multicolumn{1}{c}{\textbf{Elvia}}\\
Industry           &       SARIMAX (1,0,2)$\times$(1,1,1)24\\
Household          &          SARIMAX (4,1,1)$\times$(1,1,1)24          \\
Cabin              &           SARIMAX (2,1,1)$\times$(1,1,2)24      \\
\hline
\multicolumn{1}{c}{}&\multicolumn{1}{c}{\textbf{Solar}}\\
           &       SARIMAX (1,0,2)$\times$(1,1,1)24\\
\hline
\multicolumn{1}{c}{}&\multicolumn{1}{c}{\textbf{Wind}}\\
           &       SARIMAX (4,0,1)$\times$(0,0,0)24\\
\hline
\multicolumn{1}{c}{}&\multicolumn{1}{c}{\textbf{Temperature}}\\
           &       SARIMAX (2,1,1)$\times$(1,1,1)24\\
\bottomrule
\end{tabular}
\label{tab:model_par_SARIMA}
\end{table}

A single SARIMA model is fitted using the training dataset and, as the days in the test dataset are predicted, the previous actual observations for the test days and the validation data are made available for the model using the statsmodels \texttt{append} function. The \texttt{append} function stores the results for all training observations and extends the historical observations available to the model without refitting the model parameters.

% \end{document}

\bibliographystyle{IEEEtran}
\bibliography{bibliography.bib}

\end{document}